\def\year{2020}\relax
\documentclass[letterpaper]{article} 
\usepackage{aaai20}  
\usepackage{times}  
\usepackage{helvet} 
\usepackage{courier}  
\usepackage[hyphens]{url}  
\usepackage{graphicx} 
\usepackage{graphicx}  
\usepackage{amsfonts}
\usepackage{amsmath}
\usepackage{multirow}
\usepackage{subfig}
\usepackage{booktabs} 

\title{iFAN: Image-Instance Full Alignment Networks for Adaptive Object Detection}
\author{\textbf{Chenfan Zhuang, Xintong Han, Weilin Huang\thanks{Corresponding author: whuang@malong.com}, Matthew R. Scott
}\\ 
Malong Technologies, Shenzhen, China\\
Shenzhen Malong Artificial Intelligence Research Center, Shenzhen, China\\ 
\{fan,xinhan,whuang,mscott\}@malong.com 
}
 
\begin{document}

\maketitle

\begin{abstract}
Training an object detector on a data-rich domain and applying it to a data-poor one with limited performance drop is highly attractive in industry, because it saves huge annotation cost. Recent research on unsupervised domain adaptive object detection has verified that aligning data distributions between source and target images through adversarial learning is very useful. The key is when, where and how to use it to achieve best practice. We propose Image-Instance Full Alignment Networks  (\textbf{iFAN}) to tackle this problem by precisely aligning feature distributions on both image and instance levels: 1) Image-level alignment: multi-scale features are roughly aligned by training adversarial domain classifiers in a hierarchically-nested fashion. 2) Full instance-level alignment: deep semantic information and elaborate instance representations are fully exploited to establish a strong relationship among categories and domains. Establishing these correlations is formulated as a metric learning problem by carefully constructing instance pairs. Above-mentioned adaptations can be integrated into an object detector (e.g.\ Faster R-CNN), resulting in an end-to-end trainable framework where multiple alignments can work collaboratively in a coarse-to-fine manner. In two domain adaptation tasks: synthetic-to-real (SIM10K $\rightarrow$ Cityscapes) and normal-to-foggy weather  (Cityscapes $\rightarrow$ Foggy Cityscapes), iFAN outperforms the state-of-the-art methods with a boost of $10\%+$ AP over the source-only baseline.
\end{abstract}

\section{Introduction}

\begin{figure*}
\centering
\includegraphics[width=0.85\textwidth]{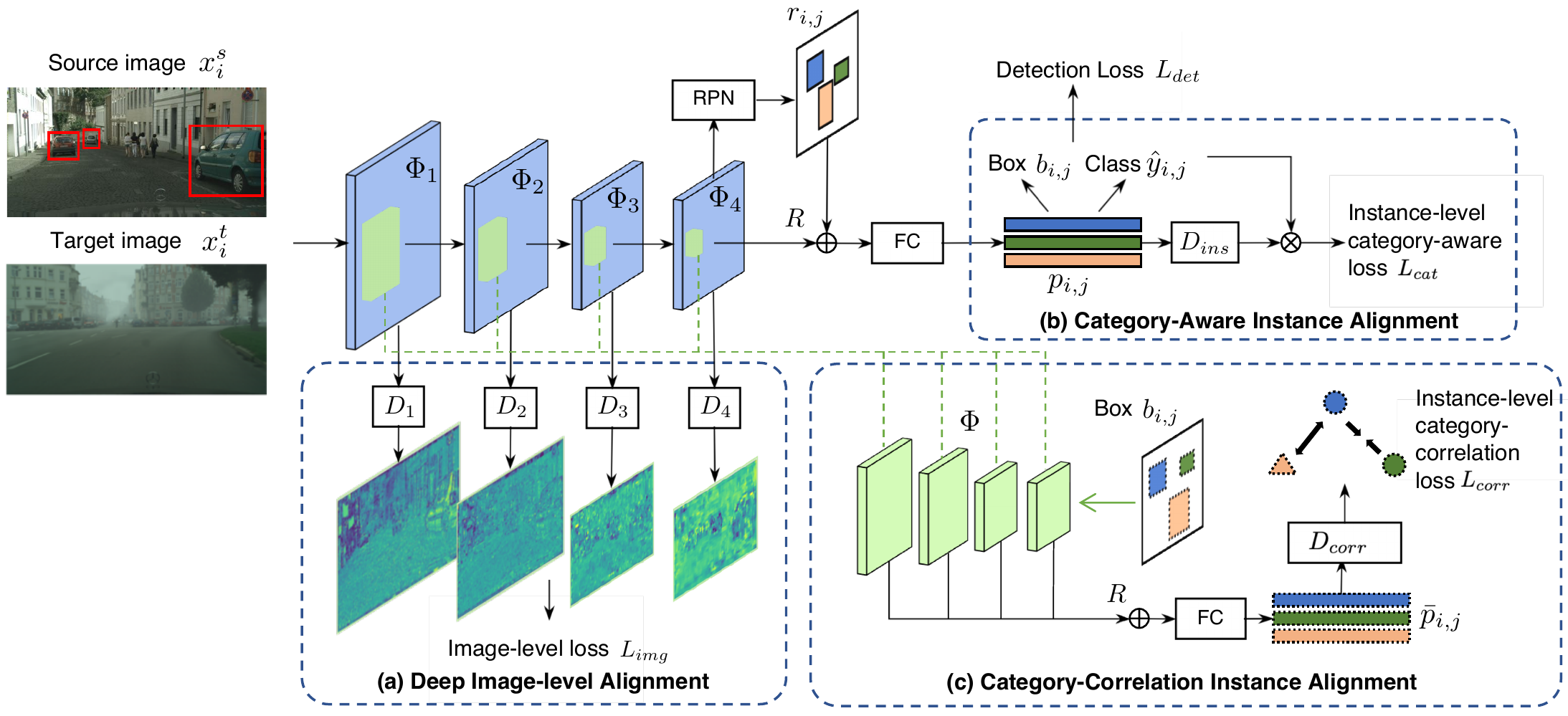}
\caption{\textbf{Overview of the proposed iFAN with a Faster R-CNN detector}.  (a) \textbf{Image-level Adaptation}: Hierarchical domain classifiers align image-level features at different semantic levels. (b) \textbf{Category-Aware Instance Adaptation}: ROI-pooled instance features are aligned in a category-aware fashion guided by the corresponding predicted class labels. (c) \textbf{Category-Correlation Instance Adaptation}: The predicted bounding boxes are utilized to extract accurate representations for the instances, then paired as input to a metric learning framework to learn the correlations across domains and categories.}
\label{fig:overview}
\end{figure*}

Training neural networks on one domain that generalizes well on another domain can significantly reduce the cost for human labeling, making domain adaptation a hot research topic. Researchers have studied the effectiveness of domain adaptation in various tasks, including image classification \cite{kumar2018co,saito2017asymmetric,shu2018dirt,long2018conditional,long2017deep}, object detection \cite{Chen2018Domain,Saito2018Strong,wang2019few,roychowdhury2019automatic,cai2019exploring,zhu2019adapting} and semantic segmentation \cite{wu2018dcan,zhang2017curriculum,hoffman2018cycada}.  In this paper, we aim to train a high-performance unsupervised domain adaptive object detector on a fully-annotated source domain and  apply it to an unlabeled target domain. For example, an object detector, trained on synthesized images generated from a game engine such as SIM10K \cite{Johnson2016Driving} where object bounding boxes are readily available, can be applied to real-world images from a target domain, such as Cityscapes \cite{cordts2016cityscapes} or KITTI \cite{geiger2013vision}.

Recently, many efforts have been devoted into developing cross-domain models with unsupervised domain adaption. Existing approaches mainly focus on aligning deep features directly between source domain and target domain.
In the context of object detection, the alignment is usually achieved by domain-adversarial training \cite{ganin2014unsupervised,tzeng2017adversarial} at different stages of object detectors. For example, based on a Faster R-CNN framework \cite{ren2015faster}, \cite{Chen2018Domain} aligned the feature maps in backbone with an image-level adaptation module; then aligned the ROI-pooled features before feeding them into the final classifier and box regressor. \cite{Saito2018Strong} strongly aligned patch distributions of the low-level features (e.g.\  \texttt{conv3} layer) to enhance local consistence and weakly aligned the global image-level features before RPN. 

We follow this line of research to develop multi-level domain alignments for cross-domain object detection, as shown in Figure \ref{fig:overview}. Unlike previous approaches that merely concern image-level alignment at a single convolutional layer (e.g.\ \cite{wang2019few,Chen2018Domain}), we design hierarchically-nested domain discriminators to reduce domain discrepancies in accordance with various characteristics in the network hierarchies (Figure \ref{fig:overview}a); meanwhile the instance-level features are carefully aligned, making use of the ROI-level representations. Notice that the traditional instance-level alignments, such as \cite{Chen2018Domain}, attempt to learn domain-invariant features without fully  exploring the semantic category-level information. This inevitably leads to a performance drop due to the misalignment of objects within the same category. To address this problem, we develop a category-aware instance-level adaptation by leveraging object classification results of the detector (Figure \ref{fig:overview}b). Finally, we propose a novel category-correlation instance alignment: utilize the predicted bounding boxes to attain the refined instance representations and precisely align them using deep metric learning - establish the correlations among domains and categories, as shown in Figure  \ref{fig:overview}c.

\noindent \textbf{Contributions.} The main contributions of this work are four-fold: 1) To mitigate the domain shift occurred in multiple semantic levels, we apply domain-adversarial training on multiple intermediate layers, allowing us to align multi-level features more effectively;
2) A category-aware instance-level alignment is then proposed to align ROI-based features, subtly incorporating deep category information;
3) We formulate the category-correlation instance-level alignment to a metric learning problem, further study the cross-domain category correlations;
4) Our approach surpasses the state-of-the-art unsupervised domain adaptive object detectors, e.g.\ \cite{wang2019few,Chen2018Domain,Saito2018Strong} on synthetic-to-real (SIM10K $\rightarrow$ Cityscapes) and normal-to-foggy weather (Cityscapes $\rightarrow$ Foggy Cityscapes) tasks.

\section{Related Work}

\noindent\textbf{Unsupervised Domain Alignment.}
Unsupervised domain adaptation (UDA) refers to train domain invariant models on images with annotated images from source domain and images from target domain without any annotation. Many UDA methods show the effectiveness of distribution matching by reducing the domain gap. \cite{saito2017maximum} focused on generating features to minimize the discrepancies between two classifiers which are trained to maximize the discrepancies on target samples. \cite{TAT_2019_ICML} generated transferable
examples to fill in the gap between the source and
target domain by adversarially training deep
classifiers to output consistent predictions over the transferable examples. However, since object detectors often generate numerous region proposals, many of which could be background or beyond the given classes, these methods can not fit very well in object detection. Another series of solutions \cite{wu2018dcan,inoue2018cross,hoffman2018cycada}, following the success of unsupervised image-to-image translation networks \cite{zhu2017unpaired,liu2017unsupervised,huang2017adain}, directly aligned pixel-level distributions by transferring source images into the target style, and then trained models on the transferred images. Inspired by Generative Adversarial Networks \cite{goodfellow2014generative}, training a domain discriminator to identify the source from the target and then reacting on the feature extractor to deceive the discriminator, has been frequently used and proven efficient \cite{ganin2014unsupervised,Saito2018Strong,Chen2018Domain,long2018conditional,tzeng2017adversarial,liu2016coupled}.

\noindent\textbf{Adaptive Object Detection.}
Object detectors with deep architectures \cite{girshick2015fast,girshick2014R-CNN,ren2015faster} play an important role in myriad computer vision applications. To eliminate the dataset bias, a number of methods have been developed to UDA object detection problems \cite{inoue2018cross,Chen2018Domain,Saito2018Strong,cai2019exploring,zhu2019adapting}. \cite{inoue2018cross} sequentially fine-tuned an object detector with an image-to-image domain transfer and weakly supervised pseudo-labeling. \cite{Chen2018Domain} developed a Faster R-CNN \cite{ren2015faster} detector with feature alignment on both image-level and instance-level. Along this direction, \cite{Saito2018Strong} forced the image-level features to be strongly aligned in the lower layers and weakly aligned in the higher layers and concatenate them together. \cite{zhu2019adapting} grouped instances into discriminatory regions and aligned region-level features across domains. \cite{cai2019exploring} learned relation graphs, which regularizes the teacher and student models to learn consistent features. Among these methods, inherent feature hierarchies and deep semantic information are not  exhaustively exploited, which motivates our method.

\noindent\textbf{Metric Learning.}
Our method also relates to metric learning aims to \cite{chopra2005learning,schroff2015facenet,oh2016deep,wang2019multi} as we construct pairs as input to the category-correlation adaptation. Metric learning approaches learn an embedding space where similar samples are pulled closer, while dissimilar ones are pushed apart from each other. In this paper, we train a metric learning model to draw two instances closer if they share the same category, or push them apart otherwise despite domains. The idea of learning with paired samples has also been utilized in few-shot domain adaptation approaches \cite{wang2019few,motiian2017few} for handling scarce annotated target data. Unlike these approaches, our category-correlation adaptation works without any supervision from the target domain. Instead, we use the predictions of classifiers as pseudo labels.

\section{Image-Instance Full Alignment Networks}

The whole pipeline of our proposed iFAN is presented in Figure \ref{fig:overview}. Given images with annotated bounding boxes from the source domain and unlabeled images from the target, our goal is to align the distributions of two domains via image-level (Figure \ref{fig:overview}a) and full instance-level alignments, including category-aware (Figure \ref{fig:overview}b) and category-correlation (Figure \ref{fig:overview}c), step by step, to boost the performance of a detector, without charging anything extra on inference. Formally, let $\{x_i^s, y_i^s\}_{i\in[N_s]}$ denote a set of  $N_s$ images $x_i^s \in \mathbb R^{H \times W \times 3}$ from the source domain, with corresponding annotations $y_i^s$. For the target domain, we only have images $\{x_i^t\}_{i\in [N_t]}$ without any annotation. An object detector (e.g.\  Faster R-CNN \cite{ren2015faster} in this paper) can be trained in the source domain by minimizing:

\begin{equation}
  L_{det} = \frac{1}{N_s}\sum_{i=1}^{N_s}L(x_i,y_i),
\end{equation}
where $L$ is the loss function for object detection. Generally, such a detector is difficult to generalize well to a new target domain due to the large domain gap.

\subsection{Deep Image-Level Alignment}
\label{sec:image-level}

Recent domain adaptive object detectors \cite{Chen2018Domain,shan2018pixel,wang2019few} commonly align image-level features to minimize the effect of domain shift, by applying a patch-based domain classifier on the intermediate features drawn from a  single convolutional layer (typically the global features before the RPN). Since the receptive field of each activation corresponds to a patch of the input image, a domain classifier can be trained to guide the networks to learn a domain-invariant representation for the image patch, and thus reduces the global image domain shift (e.g.\  image style, illumination, texture, etc.). This patch-based discriminator has also been proved to be effective on cross-domain image-to-image translation task \cite{pix2pix2016,zhu2017unpaired}.

Nevertheless, these methods just focus on features extracted from a certain layer; they may  miss the rich domain information contained in other intermediate layers, such as, the domain displacement of different scales.
Recent works of object detection \cite{lin2017feature} and image synthesis \cite{Zhang2018Photographic,Yang2019Learning}, which explore the inherent multi-scale pyramidal hierarchy of convolutional neural networks to achieve meaningful deep multi-scale representations, have greatly inspired our work.

We propose to build a hierarchically-nested domain classifier bank at the multi-scale intermediate layers, as shown in Figure \ref{fig:overview}a. Let $\Phi$ denote the backbone of an object detector. For the feature maps $\Phi_l \in \mathbb R^{H_l\times W_l \times C_l}$ from the $l^{th}$ intermediate layer, a domain classifier $D_l$ is constructed in a fully convolutional fashion (e.g.\  3 convolutional layers with $1\times1$ kernels) to distinguish source (domain label = 0) and target (domain label = 1) samples, by minimizing a mean square-error loss as \cite{Saito2018Strong,zhu2017unpaired}:

\begin{equation}
\begin{split}
L_{D_l} = \frac{1}{N_sH_lW_l}\sum_{i=1}^{N_s}\sum_{m=1}^{H_lW_l}D_l(\Phi_l(x^s_i))_m^2 + \\
\frac{1}{N_tH_lW_l}\sum_{i=1}^{N_t}\sum_{m=1}^{H_lW_l}(1 - D_l(\Phi_l(x^t_i)))_m^2. \\
\end{split}
\end{equation}

In this paper, we use \texttt{pool2}, \texttt{pool3}, \texttt{pool4}, \texttt{relu5\_3} in VGG-16 backbone \cite{simonyan2014very} or  \texttt{res2c\_relu}, \texttt{res3d\_relu}, \texttt{res4f\_relu}, \texttt{res5c\_relu} for ResNet50 \cite{he2015deep}  as the intermediate layers. Then the loss for our hierarchically-nested domain classifier bank forms:

\begin{equation}
  L_{img} = \sum^{4}_{l=1}\lambda_l L_{D_l},
  \label{eq:img}
\end{equation}

where $\lambda_l$ denotes a balancing weight, empirically set to reconcile each penalty. By minimizing $L_{img}$, the domain classifiers are forced to discriminate the multi-scale features of the source domain from the target; meanwhile, the detector is trying to generate “domain-ambiguous” features to deceive these domain discriminators via reversal gradient \cite{ganin2014unsupervised} , yielding domain-invariant features that generalize well to the target domain.

Compared to the single global domain classifier developed in \cite{Chen2018Domain,shan2018pixel}, our hierarchically-nested image-level alignment enjoys the following merits:
1) Receptive field with various sizes on the hierarchy enable the model to align image patches in a bigger range of scales in the spirit of feature pyramid network \cite{lin2017feature}.

2) Our multi-layer alignment is designed to capture multi-granularity characteristics of domains at a time, from low-level features (e.g.\  texture and color)  to high-level (e.g.\  shape). This voracious strategy can effectively reduce domain discrepancies of various kind.

3) Unlike existing domain-adversarial frameworks which might suffer from unstable training \cite{Saito2018Strong,wang2019few}, our proposed hierarchical supervisions could guide the alignment gradually and moderately,  from shallow to deep, leading to better convergence.

\subsection{Full Instance-Level Alignment}
\label{sec:instance-level}

\subsubsection{Category-Agnostic Instance Alignment}
\label{sec:category-agnostic}

Recent adaptive object detectors, e.g.\ \cite{Chen2018Domain}, also integrate a category-agnostic instance domain classifier $D_{ins}$ on the top of ROI-based features to mitigate domain shift between local instances, e.g.\  the appearance and the shape of objects. Following this line, we extend the image-level alignment to instance level. Let $p = R(f, r)$ denote the output of ROI-Align operation \cite{he2017mask}, conditioned on feature maps $f$ and a region proposal $r$. $p_{i,j} = R(\Phi_4(x_i), r_{i,j})$ is the instance feature of the $j^{th}$ region proposal of image $x_i$, as shown in Figure \ref{fig:overview}b. Our loss function of a naive instance alignment formulates:

\begin{equation}
\begin{aligned}
  L_{ins} = &\frac{1}{N_s}\sum_{i=1}^{N_s}\frac{1}{N^s_{i}}\sum_{j=1}^{N^s_{i}}D_{ins}(R(\Phi_4(x^s_i), r^s_{i,j}))^2 + \\ & \frac{1}{N_t}\sum_{i=1}^{N_t}\frac{1}{N^t_{i}}\sum_{j=1}^{N^t_i}(1 - D_{ins}(R(\Phi_4(x^t_i), r^t_{i,j})))^2,
\end{aligned}
\end{equation}
where $N^s_i$ and $N^t_i$ denote the numbers of instances in $x^s_i$ and $x^t_i$, respectively. To clarify the effectiveness of instance-level adaptation, we simply adopt the same architecture used in the image-level alignment, which consists of $1 \times 1$ convolutions. 

However, we observed that applying such an instance-level alignment from the beginning of training may not be the best practice. At early stage of the training, the predictions of detector for both source and target images are inaccurate. With the supervision of ground-truth, knowledge from the source data can be steadily learned and simultaneously transferred to the target data.
Intuitively, aligning nonsensical patches instead of the valid instances could bring negative effects: the instance alignment can make positive impact only when the detector becomes relatively stable in both source and target domains. This challenge was discussed and validated in \cite{Saito2018Strong} as well. To tackle this problem, we propose a technique called “late launch”: an activate instance-level alignment at one third of the whole training iterations. Note that the total of training iterations remains unchanged.

\subsubsection{Category-Aware Instance Alignment}
\label{sec:category-Ware}

Our image-level and category-agnostic instance-level alignments are able to blend the features from two domains together. However, category information has not been taken into consideration, and instances from two different domains are possible to be aligned incorrectly into different classes.
 For example, the feature of a car in the source may be aligned with a bus in the target, resulting in undesired performance drop.

To this end,  we propose to incorporate category information into instance-level alignment by modifying $D_{ins}$ to $C$-way output instead of the original single-way. In other words, each category owns a domain discriminator. Thus the $c^{th}$ dimension of $D_{ins}(p_{i,j})$ indicates a domain label (source = 0 or target = 1) for the instance (with corresponding features $p_{i,j}$) from category $c$.

However, category labels for the instances from the target domain are not provided; thus  the methodology to assign the category labels to target proposals is pending. Enlightened by the pseudo-labeling approach described in \cite{inoue2018cross}, we directly use the classifier output of the detector $\hat y_{i,j}$ as \textit{soft pseudo-labels} for target instances, as shown in Figure \ref{fig:overview}b. The classifier output indicates the probability distribution of how likely an instance belongs to the $C$ classes. According to the possibility, domain classifiers of each category independently update their own parameters.
As a result, the category-aware instance alignment loss takes the following form:

\begin{equation}
\begin{aligned}
  L_{cat} = & \frac{1}{N_s}\sum_{i=1}^{N_s}\frac{1}{N^s_{i}}\sum_{j=1}^{N^s_{i}}\sum_{c=1}^{C}\hat y^s_{i,j,c} D_{ins}(p^s_{i,j})_c^2 + \\ & \frac{1}{N_t}\sum_{i=1}^{N_t}\frac{1}{N^t_{i}}\sum_{j=1}^{N^t_i}\sum_{c=1}^{C}\hat y^t_{i,j,c}(1 - D_{ins}(p^t_{i,j})_c)^2,
\end{aligned}
\end{equation}
where the loss of domain classifier is weighted by the predicted category probability. Notice that, for source instances, we use the predicted labels in the same way as we found that this soft assignment policy factually works better than using ground truth.

\subsubsection{Category-Correlation Instance Alignment}
\label{sec:metric-level}

Due to the location misalignment between a coarse region proposal and its accurate bounding box, ROI-based features may fail to precisely characterize the instances. Popular object detectors prefer to refine the bounding boxes in an iterative \cite{gidaris2015object} or cascaded \cite{cai2018cascade} manner to reduce such misalignment for higher accuracy. Similarly, in this paper, we propose to enhance instance representations by mapping the predicted bounding boxes back to the backbone feature maps, and crop the selective features out for further alignment. This process is illustrated in Figure  \ref{fig:overview}c. Formally, for the $j^{th}$ predicted object in image $x_i$, we use $b_{i,j}$ to denote its predicted bounding box, then pool the corresponding feature maps $\Phi_l(x_i)$  at the $l^{th}$ layer with ROI-Align \cite{he2017mask}, finally shaping a group of representations for this instance: $p_{i,j,l} = R(\Phi_l(x_i), b_{i,j})$.

Moreover, following the principle of image-level alignment, we fuse these representations to combine all possible information together.
The feature maps $p_{i,j,l}$ ($l = 1,2,3,4$) are individually passed into $1\times1$ convolutions to generate features with 256 channels; then element-wise summation is applied, yielding the refined features $\overline p_{i,j}$. 
Compared with the ROI-pooled features computed from a  single layer  ($\Phi_1 \sim \Phi_4$), the summation operation can improve $0.5\%+$ mAP.

We then project the refined instance features into an embedding space through a fully-connected layer denoted as $D_{corr}$. Given a pair of instances $f_i$ and $f_j$, it should belong to one of the four groups according to its domain and category:  1) same-domain and same-category $\mathcal S_{sdsc}$;  2) same-domain and different-category $\mathcal S_{sddc}$;  3) different-domain and same-category $\mathcal S_{ddsc}$;  4) different-domain and different-category $\mathcal S_{dddc}$. We found that  minimizing the distances in $S_{sdsc}$ and maximizing in $S_{dddc}$ are two simplistic tasks, thanks to the previous alignments in the object detector. Therefore we only focus on $S_{sddc}$ and $S_{ddsc}$ to optimize the correlations of domains and categories via metric learning.

With $D_{corr}$ used as a \textit{metric discriminator}, we can minimize the following contrastive loss \cite{chopra2005learning}:

\begin{equation}
\begin{aligned}
  L_{corr} = & \frac{1}{|\mathcal S_{sddc}|}\sum_{(f_i, f_j)\in S_{sddc}}d(f_i, f_j)^2 + \\
  & \frac{1}{|\mathcal S_{ddsc}|}\sum_{(f_i, f_j)\in S_{ddsc}}\max(0, m - d(f_i, f_j))^2,
\end{aligned}
\end{equation}

where $d(f_i,f_j) = ||D_{corr}(f_i) - D_{corr}(f_j)||_2$ denotes the Euclidean distance, and $m$ is a fixed margin. Remember that we are under an adversarial training; hence, $D_{corr}$ ought to pull together the instance pairs in the same domain even from different categories, while pushing apart the pairs from different domains but of the same category. On the contrary, the Faster R-CNN is trying to confuse this metric discriminator by maximizing $L_{corr}$. As a result, the object detector can generate features that encourage: 1) Different categories are well separated within the same domain ($S_{sddc}$); 2) Features are domain-invariant for instances of the same class ($S_{ddsc}$). Both of them are the desired properties for an ideal domain adaptive classifier.

Since the category labels of target instances are not available, we again use the predicted labels of the instances to construct pairs. Similarly, “late launch” technique is used here too, as mentioned in the category-aware instance alignment.

\subsection{Training and Inference}

The full training objective of our method is:

\begin{equation}
\begin{aligned}
  \min_{G} \max_{D} & L_{det}(G) - \\& \lambda_{adv}(L_{img}(G,D) +  L_{cat}(G,D)  + L_{corr}(G,D)),
  \label{eq:obj}
\end{aligned}
\end{equation}

where $G$ denotes a Faster R-CNN object detector, and $D$ indicates one of the three domain classifiers: $D_{l}$, $D_{cat}$, or $D_{corr}$.
 $\lambda_{adv}$ is the weight of adversarial loss to balance the penalty between the detection and adaptation task. The minimax loss function  is implemented by a gradient reverse layer (GRL) \cite{ganin2014unsupervised}.

No worries to increase the burden on inference stage, because all alignments are only carried out during training and the modules can be easily peeled off.   Although iFAN increases the computational cost during training, luckily not too much, the inference speed is identical to a vanilla Faster R-CNN as \cite{Chen2018Domain,wang2019few,cai2019exploring} , and is faster than \cite{Saito2018Strong}, whose inference involves computing the outputs of domain classifiers.

\section{Experiments and Results}

\subsection{Experimental Setup}

\begin{table}[t!]
\centering
\resizebox{1.0\columnwidth}{!}{
\begin{tabular}{l|c|c|c}
\toprule
Method & Backbone & S $\rightarrow$ C &  C $\rightarrow$ F\\\hline
& \\[\dimexpr-\normalbaselineskip+2pt]

Oracle & VGG16 & 61.1 & 38.9 \\\hline
& \\[\dimexpr-\normalbaselineskip+2pt]

Source-only Faster R-CNN & VGG16 & 34.9 & 16.9 \\
ADDA \cite{tzeng2017adversarial} & VGG16 & 36.1 & 24.9 \\
DT \cite{zhu2017unpaired} + FT & VGG16 & 36.8 & 26.1 \\

DA-Faster \cite{Chen2018Domain} & VGG16 & 40.0 & 27.6 \\
SW \cite{Saito2018Strong} &  VGG16 & 40.1 & 34.3 \\
Few-shot \cite{wang2019few} & VGG16 & 41.2 & 31.3 \\
SelectAlign \cite{zhu2019adapting}  & VGG16 & 43.0 &  33.8 \\
\textbf{iFAN} & VGG16 &\textbf{46.9} & \textbf{35.3} \\\hline
& \\[\dimexpr-\normalbaselineskip+2pt]

Oracle & ResNet50 & 66.4 & 45.2 \\\hline
& \\[\dimexpr-\normalbaselineskip+2pt]

Source-only Faster R-CNN & ResNet50 & 35.1 & 21.0 \\
MTOR \cite{cai2019exploring} & ResNet50 & 46.6 & 35.1 \\
\textbf{iFAN} & ResNet50 &\textbf{47.1} & \textbf{36.2} \\

\bottomrule
\end{tabular}
}
\caption{Comparison with other methods. Mean average precision (mAP, $\%$) on SIM10K $\rightarrow$ Cityscapes  (S $\rightarrow$ C) and Cityscapes $\rightarrow$ Foggy Cityscapes (C $\rightarrow$ F).}
\label{tab:sim10k}
\end{table}

\begin{table}[t!]
\centering
\begin{tabular}{c|ccc|cc}
\toprule
& img & ins & corr & AP & gain\\\hline
Source only&&&& 34.9 & - \\\hline
\multirow{4}{*}{iFAN}&$\checkmark$  &&& 43.0 & 8.1 \\
&$\checkmark$  &$\checkmark$ && 46.1 & 11.2 \\
&$\checkmark$  & &$\checkmark$ & 45.3 & 10.4 \\
&$\checkmark$  &$\checkmark$ &$\checkmark$ & \textbf{46.9} & \textbf{12.0} \\
\bottomrule
\end{tabular}

\caption{Ablations on  SIM10K $\rightarrow$ Cityscapes. img, ins, corr denote our image-level, category-agnostic and category-correlation instance alignment respectively. No category-aware alignment in this scenario, since only “car” is evaluated. }
\label{tab:sim10k_ablation}
\end{table}

\begin{table*}[t!]
\centering
\begin{tabular}{c|cccc|cccccccc|cc}
\toprule
& img & ins & cat & corr & person	& car &	moto & rider & bicycle	& bus & train & truck & mAP & gain \\\hline
Source only&&&&& 21.5&28.8&13.6&21.9&21.4&16.0&5.0&7.0&16.9& - \\\hline
\multirow{5}{*}{iFAN}&$\checkmark$  &&&&\textbf{33.0}&47.2&25.2&\textbf{41.3}&\textbf{33.3}&41.1&15.2&23.6&32.5& 15.6 \\
&$\checkmark$  &$\checkmark$ & &&32.3&48.4&\textbf{28.1}&41.0&32.7&41.4&23.0&22.6&33.1& 16.2 \\
&$\checkmark$  &&$\checkmark$ &&32.4&\textbf{48.9}&23.9&38.3&32.5&44.8&28.5&27.5&34.6& 17.7 \\
&$\checkmark$  && &$\checkmark$ &32.3&47.8&20.5&38.6&32.9&43.5&\textbf{33.0}&27.3&34.5&17.6 \\
&$\checkmark$  &&$\checkmark$ &$\checkmark$ &32.6&48.5&22.8&40.0&33.0&\textbf{45.5}&31.7&\textbf{27.9}&\textbf{35.3}&\textbf{18.4} \\
\bottomrule
\end{tabular}
\caption{Ablations on  Cityscapes $\rightarrow$ Foggy Cityscapes. img, ins, cat and corr denote our image-level, category-agnostic, category-aware and category-correlation instance alignment respectively. }
\label{tab:foggy_ablation}
\end{table*}

\textbf{Datasets} We evaluate \textbf{iFAN} on two domain adaptation scenarios: 1) train on SIM10K \cite{Johnson2016Driving} and test on Cityscapes \cite{cordts2016cityscapes} dataset (SIM10K $\rightarrow$ Cityscapes); 2) train on Cityscapes \cite{cordts2016cityscapes} and test on Foggy Cityscapes \cite{sakaridis2018semantic} (Cityscapes $\rightarrow$ Foggy). Rendered by the “Grand Theft Auto” game engine, the SIM10K dataset consists of 10,000 images with 58,701 bounding boxes annotated for cars. The Cityscapes dataset has 3,475 images of 8 object categories taken from real urban scenes, where 2,975 images are used for training and the remaining 500 for evaluation. We follow \cite{Saito2018Strong,Chen2018Domain} to extract bounding box annotations by taking the tightest rectangles of the instance masks. The Foggy Cityscapes \cite{sakaridis2018semantic} dataset was created by applying fog synthesis on the Cityscapes dataset and inherit the annotations. In the SIM10K $\rightarrow$ Cityscapes scenario, only the car category is used for training and evaluation, while for Cityscapes $\rightarrow$ Foggy, all 8 categories are considered. We use an average precision with threshold $=$ 0.5 ($mAP_{50}$) as the evaluation metric for object detection.

\noindent\textbf{Implementation Details.}
To make a fair comparison with existing approaches, we strictly follow the implementation details of \cite{Saito2018Strong}. We adopt Faster R-CNN \cite{ren2015faster} + ROI-alignment \cite{he2017mask} and implement all with maskrcnn-benchmark \cite{massa2018mrcnn}. The shorter side of training and test images are set to 600. The detector is first trained with a learning rate of $lr=0.001$ for 50K iterations, and then $lr=0.0001$ for another 20K iterations. The category-agnostic/aware instance-level alignment \textit{late launches} at 30K-th iteration and category-correlation alignment at 50K-th. The \textit{late launches} timing is empirically set according to the loss curve: a new alignment starts when the previous ones go stable. We set $\lambda_{adv}=0.1$ in Eqn. \ref{eq:obj} and $\lambda_{l}=1.0$ in Eqn. \ref{eq:img}. The embedding dimension of category-correlation alignment is set to 256, with a margin of $m = 1.0$. VGG-16 is used as the backbone if not specifically indicated.

\noindent\textbf{Competing Methods.}
We compare \textbf{iFAN} with the following baselines and recent state-of-the-art methods:
1) Faster R-CNN \cite{ren2015faster}: a vanilla Faster R-CNN  trained only on the source domain.
2) ADDA \cite{tzeng2017adversarial}: the deep features from the last layer of the detector backbone are aligned with a global domain classifier.
3) Domain Transfer + Fine-Tuning (DT \cite{zhu2017unpaired}+FT): a CycleGAN \cite{zhu2017unpaired} is used to transfer the source images to the target style, and then a Faster R-CNN detector is trained on the transferred images. A similar approach is also described in \cite{inoue2018cross}.
4) Domain adaptive Faster R-CNN (DA-Faster) \cite{Chen2018Domain}: an image-level domain classifier used to align global features, an instance domain classifier for aligning the instance representations and a consistency loss for regularizing the image-level and instance-level loss to consistency.
5) Strong-Weak Alignment (SW) \cite{Saito2018Strong}: strong local alignment on the top of \texttt{conv3\_3} and weak global alignment  on \texttt{relu5\_3}. The outputs of the alignment modules are later concatenated to the instance features which are then fed into the classifier and box regressor.
6) Few-shot adaptive Faster R-CNN (Few-shot) \cite{wang2019few}: multi-scale local features are paired for image-level alignment, with semantic instance-level alignment of  object features.
7) Selective cross-domain alignment (SelectAlign) \cite{zhu2019adapting}: discover the discriminatory regions by clustering instances and align two domains at the region level.
8) Mean Teacher with Object Relations (MTOR) \cite{cai2019exploring}: capture the object relations between teacher and student models by proposing graph-based consistency losses, with 50-layer ResNet \cite{he2015deep} as backbone (we follow its implementation details for comparison).

\subsection{Main Results}

\begin{figure*}
    \centering
    \includegraphics[width=0.8\textwidth]{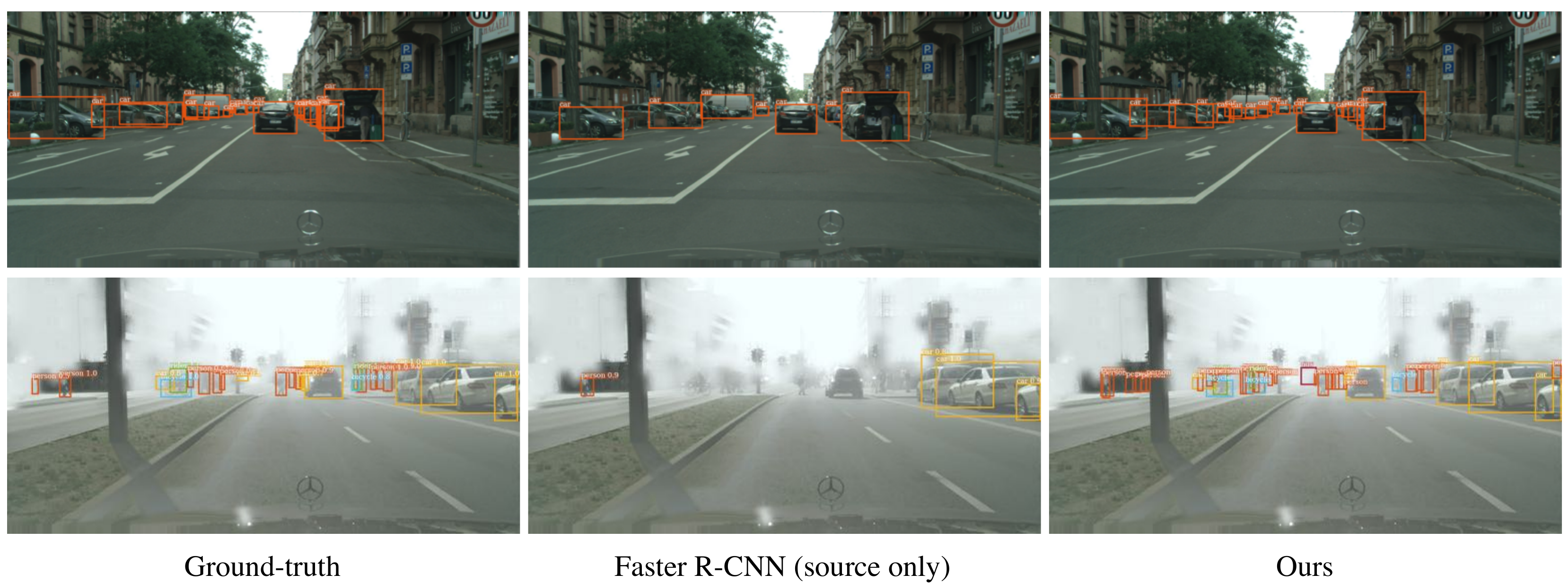}
    \caption{Qualitative results. Top: SIM10K $\rightarrow$ Cityscapes. Bottom: Cityscapes $\rightarrow$ Foggy Cityscapes.}
    \label{fig:vis}
\end{figure*}

\begin{figure}[h!]
\centering
    \includegraphics[width=.43\textwidth]{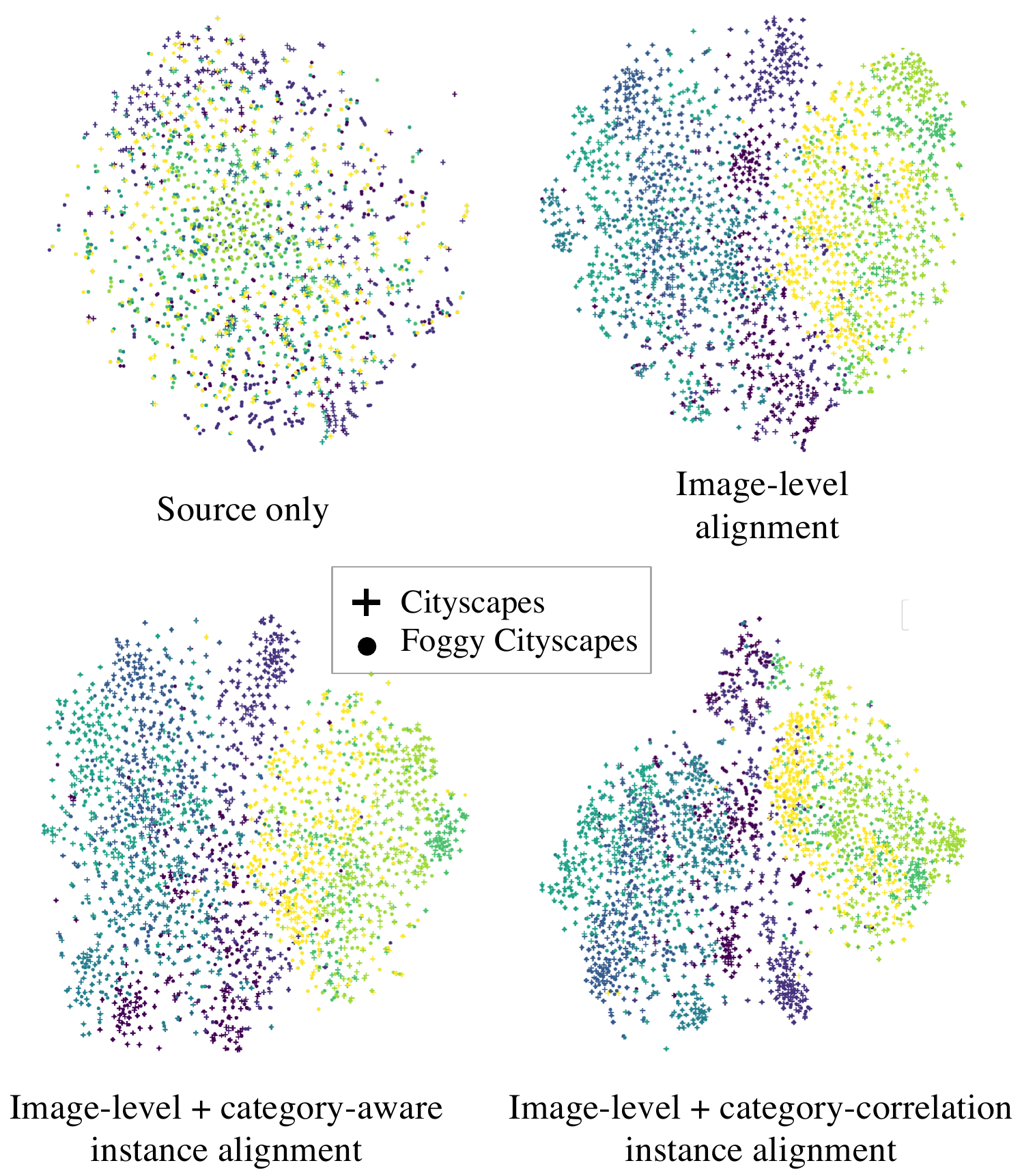}
    \caption{Visualization of ROI features from iFAN trained on Cityscapes $\rightarrow$ Foggy Cityscapes. Colors represent categories. We can see that intra- and inter-class relations are gradually optimized when deeper semantic information is encoded.}
\label{fig:tsne}
\end{figure}

Our method is compared with state-of-the-art UDA object detectors in Table \ref{tab:sim10k}. As can be found, all methods can improve the performance of baseline (Faster R-CNN trained only on the source domain) by learning domain-invariant features at various stages in the networks. Particularly, in the SIM10K $\rightarrow$ Cityscapes scenario, our method obtains more than $10\%$ AP improvement ($34.9\%$ $\rightarrow$ $46.9\%$) over the source model, achieving a higher accuracy than state-of-the-art:  $46.0\%$ (iFAN) vs $43.0\%$ \cite{zhu2019adapting}. For Cityscapes $\rightarrow$ Foggy Cityscapes, our method doubles the $mAP$ of the source-only model ($16.9\%$ $\rightarrow$ $35.3\%$) with VGG16 backbone, outperforming the other approaches by at least $1\%$ on $mAP$. 

 In Figure \ref{fig:vis}, we illustrate two example results from source-only baseline and iFAN. Clearly, iFAN generalizes better to the novel data by detecting more challenging cases. Figure \ref{fig:tsne} shows that in Cityscapes $\rightarrow$ Foggy Cityscapes task, how instances move from original chaos to domain-invariant state with category cohesion, conforming the advances of iFAN.

We also report oracle results by training a Faster R-CNN detector directly on the fully-annotated training images on target domain. We can see that here still exists a performance gap between iFAN and the oracle result, especially on SIM10K $\rightarrow$ Cityscapes, which indicates that more sophisticated UDA methods are yet required to match the performance.

\subsection{Discussions}

\textbf{Ablation Study.} We conduct ablation study by isolating each component in iFAN. The results are presented in Table \ref{tab:sim10k_ablation} and \ref{tab:foggy_ablation}. Here are our observations: 1) With image-level alignment alone, we achieve a significant performance gain; 2) Instance-level alignment further reduces the domain discrepancies for objects; 3) For multi-class dataset like Cityscapes $\rightarrow$ Foggy, category-aware and category-correlation instance-level  alignments obtain a higher accuracy than category-agnostic alignment, suggesting that the exploration on richer semantic information of instances can work better. 4) Integrating deep image-level with full instance-level alignments reaches the best results.

\noindent\textbf{Layers Used in Image-Level Alignment.}
Table \ref{tab:image-level} shows $mAP$ of disparate combinations of intermediate features in image-level alignment. Our hierarchically-nested discriminators are designed for characterizing domain shift at different semantic levels, and thus yield higher performance than individual layer. Moreover, we found that the lower layers work better than the higher ones, indicating that domain discrepancies are caused more heavily by low-level features like texture, color or illumination.

\noindent\textbf{Timing for late launch.}
The instance-level alignment is activated in the middle of the training procedure. In Table \ref{tab:late}, we report $AP$@Car on SIM10K $\rightarrow$ Cityscapes with various \textit{late launch} timings for category-agnostic instance alignment. As expected, starting instance-level alignment too early causes performance degradation: $42.1\%$ (start at 10K-th iters) vs $43.0\%$ (image-level alignment only); while too late, the instance discriminators fail to fully converge. Similarly, timing of the late launch is pivotal to the joint category-aware and category-correlation alignment.

\begin{table}[t!]
\centering
\subfloat[Comparisons of different image-level alignment strategies. Multi-level features outperform individuals. \label{tab:image-level}]{
\begin{tabular}{c|c|c|c|c|c}
\toprule
Layers & $\Phi_1$ & $\Phi_2$& $\Phi_3$ & $\Phi_4$ & $\Phi_1\sim\Phi_4$\\\hline
& \\[\dimexpr-\normalbaselineskip+2pt]

AP (\%) & 41.3& 41.8& 39.4 &38.3&\textbf{43.0}\\

\bottomrule
\end{tabular}
}

\subfloat[Effect on which training iteration to start instance-level category-agnostic alignment.
\label{tab:late}]{

\begin{tabular}{c|c|c|c|c|c}
\toprule
Start Step & 10K& 20K&30K&40K& 50K\\\hline
& \\[\dimexpr-\normalbaselineskip+2pt]

AP (\%) & 42.1& 43.6&\textbf{46.1}&45.2& 45.1\\
\bottomrule
\end{tabular}
}
\caption{More results on SIM10K $\rightarrow$ Cityscapes.}
\end{table}

\section{Conclusion}
We have presented a new domain alignment framework \textbf{iFAN} for unsupervised domain adaptive object detection. Two granularity levels of alignments are introduced:  1) Image-level alignment is implemented by aggregating multi-level deep features; 2) Full instance-level alignment is at first improved by explicitly encoding category information of the instances, and then enhanced by learning cross-domain category correlations using a metric learning formulation.
The proposed iFAN achieves new state-of-the-art performance on two domain adaptive object detection tasks: synthetic-to-real (SIM10K $\rightarrow$ Cityscapes) and normal-to-foggy weather (Cityscapes $\rightarrow$ Foggy Cityscapes), with a boost of more than 10\% AP over the source-only baseline.

\begin{small}
\bibliographystyle{aaai}\bibliography{reference}

\begin{thebibliography}{}

\bibitem[\protect\citeauthoryear{Cai and Vasconcelos}{2018}]{cai2018cascade}
Cai, Z., and Vasconcelos, N.
\newblock 2018.
\newblock Cascade r-cnn: Delving into high quality object detection.
\newblock In {\em CVPR}.

\bibitem[\protect\citeauthoryear{Cai \bgroup et al\mbox.\egroup
  }{2019}]{cai2019exploring}
Cai, Q.; Pan, Y.; Ngo, C.-W.; Tian, X.; Duan, L.; and Yao, T.
\newblock 2019.
\newblock Exploring object relation in mean teacher for cross-domain detection.
\newblock In {\em CVPR}.

\bibitem[\protect\citeauthoryear{Chen \bgroup et al\mbox.\egroup
  }{2018}]{Chen2018Domain}
Chen, Y.; Li, W.; Sakaridis, C.; Dai, D.; and Van~Gool, L.
\newblock 2018.
\newblock Domain adaptive faster r-cnn for object detection in the wild.
\newblock In {\em CVPR}.

\bibitem[\protect\citeauthoryear{Chopra \bgroup et al\mbox.\egroup
  }{2005}]{chopra2005learning}
Chopra, S.; Hadsell, R.; LeCun, Y.; et~al.
\newblock 2005.
\newblock Learning a similarity metric discriminatively, with application to
  face verification.
\newblock In {\em CVPR}.

\bibitem[\protect\citeauthoryear{Cordts \bgroup et al\mbox.\egroup
  }{2016}]{cordts2016cityscapes}
Cordts, M.; Omran, M.; Ramos, S.; Rehfeld, T.; Enzweiler, M.; Benenson, R.;
  Franke, U.; Roth, S.; and Schiele, B.
\newblock 2016.
\newblock The cityscapes dataset for semantic urban scene understanding.
\newblock In {\em CVPR}.

\bibitem[\protect\citeauthoryear{Ganin and
  Lempitsky}{2015}]{ganin2014unsupervised}
Ganin, Y., and Lempitsky, V.
\newblock 2015.
\newblock Unsupervised domain adaptation by backpropagation.
\newblock In {\em ICML}.

\bibitem[\protect\citeauthoryear{Geiger \bgroup et al\mbox.\egroup
  }{2013}]{geiger2013vision}
Geiger, A.; Lenz, P.; Stiller, C.; and Urtasun, R.
\newblock 2013.
\newblock Vision meets robotics: The kitti dataset.
\newblock {\em IJRR}.

\bibitem[\protect\citeauthoryear{Gidaris and
  Komodakis}{2015}]{gidaris2015object}
Gidaris, S., and Komodakis, N.
\newblock 2015.
\newblock Object detection via a multi-region and semantic segmentation-aware
  cnn model.
\newblock In {\em ICCV}.

\bibitem[\protect\citeauthoryear{Girshick \bgroup et al\mbox.\egroup
  }{2014}]{girshick2014R-CNN}
Girshick, R.; Donahue, J.; Darrell, T.; and Malik, J.
\newblock 2014.
\newblock Rich feature hierarchies for accurate object detection and semantic
  segmentation.
\newblock In {\em CVPR}.

\bibitem[\protect\citeauthoryear{Girshick}{2015}]{girshick2015fast}
Girshick, R.
\newblock 2015.
\newblock Fast r-cnn.
\newblock In {\em ICCV}.

\bibitem[\protect\citeauthoryear{Goodfellow \bgroup et al\mbox.\egroup
  }{2014}]{goodfellow2014generative}
Goodfellow, I.; Pouget-Abadie, J.; Mirza, M.; Xu, B.; Warde-Farley, D.; Ozair,
  S.; Courville, A.; and Bengio, Y.
\newblock 2014.
\newblock Generative adversarial nets.
\newblock In {\em NeurIPS}.

\bibitem[\protect\citeauthoryear{He \bgroup et al\mbox.\egroup
  }{2016}]{he2015deep}
He, K.; Zhang, X.; Ren, S.; and Sun, J.
\newblock 2016.
\newblock Deep residual learning for image recognition.
\newblock In {\em CVPR}.

\bibitem[\protect\citeauthoryear{He \bgroup et al\mbox.\egroup
  }{2017}]{he2017mask}
He, K.; Gkioxari, G.; Doll{\'a}r, P.; and Girshick, R.
\newblock 2017.
\newblock Mask r-cnn.
\newblock In {\em ICCV}.

\bibitem[\protect\citeauthoryear{Hoffman \bgroup et al\mbox.\egroup
  }{2018}]{hoffman2018cycada}
Hoffman, J.; Tzeng, E.; Park, T.; Zhu, J.-Y.; Isola, P.; Saenko, K.; Efros, A.;
  and Darrell, T.
\newblock 2018.
\newblock Cycada: Cycle-consistent adversarial domain adaptation.
\newblock In {\em ICML}.

\bibitem[\protect\citeauthoryear{Huang and Belongie}{2017}]{huang2017adain}
Huang, X., and Belongie, S.
\newblock 2017.
\newblock Arbitrary style transfer in real-time with adaptive instance
  normalization.
\newblock In {\em ICCV}.

\bibitem[\protect\citeauthoryear{Inoue \bgroup et al\mbox.\egroup
  }{2018}]{inoue2018cross}
Inoue, N.; Furuta, R.; Yamasaki, T.; and Aizawa, K.
\newblock 2018.
\newblock Cross-domain weakly-supervised object detection through progressive
  domain adaptation.
\newblock In {\em CVPR}.

\bibitem[\protect\citeauthoryear{Isola \bgroup et al\mbox.\egroup
  }{2017}]{pix2pix2016}
Isola, P.; Zhu, J.-Y.; Zhou, T.; and Efros, A.~A.
\newblock 2017.
\newblock Image-to-image translation with conditional adversarial networks.
\newblock In {\em CVPR}.

\bibitem[\protect\citeauthoryear{Johnson-Roberson \bgroup et al\mbox.\egroup
  }{2017}]{Johnson2016Driving}
Johnson-Roberson, M.; Barto, C.; Mehta, R.; Sridhar, S.~N.; Rosaen, K.; and
  Vasudevan, R.
\newblock 2017.
\newblock Driving in the matrix: Can virtual worlds replace human-generated
  annotations for real world tasks?
\newblock In {\em ICRA}.

\bibitem[\protect\citeauthoryear{Kumar \bgroup et al\mbox.\egroup
  }{2018}]{kumar2018co}
Kumar, A.; Sattigeri, P.; Wadhawan, K.; Karlinsky, L.; Feris, R.; Freeman, B.;
  and Wornell, G.
\newblock 2018.
\newblock Co-regularized alignment for unsupervised domain adaptation.
\newblock In {\em NeurIPS}.

\bibitem[\protect\citeauthoryear{Lin \bgroup et al\mbox.\egroup
  }{2017}]{lin2017feature}
Lin, T.-Y.; Doll{\'a}r, P.; Girshick, R.~B.; He, K.; Hariharan, B.; and
  Belongie, S.~J.
\newblock 2017.
\newblock Feature pyramid networks for object detection.
\newblock In {\em CVPR}.

\bibitem[\protect\citeauthoryear{Liu and Tuzel}{2016}]{liu2016coupled}
Liu, M.-Y., and Tuzel, O.
\newblock 2016.
\newblock Coupled generative adversarial networks.
\newblock In {\em NeurIPS}.

\bibitem[\protect\citeauthoryear{Liu \bgroup et al\mbox.\egroup
  }{2019}]{TAT_2019_ICML}
Liu, H.; Long, M.; Wang, J.; and Jordan, M.~I.
\newblock 2019.
\newblock Transferable adversarial training: A general approach to adapting
  deep classifiers.
\newblock In {\em Proceedings of the 36th International Conference on Machine
  Learning}.

\bibitem[\protect\citeauthoryear{Liu, Breuel, and
  Kautz}{2017}]{liu2017unsupervised}
Liu, M.-Y.; Breuel, T.; and Kautz, J.
\newblock 2017.
\newblock Unsupervised image-to-image translation networks.
\newblock In {\em NeurIPS}.

\bibitem[\protect\citeauthoryear{Long \bgroup et al\mbox.\egroup
  }{2017}]{long2017deep}
Long, M.; Zhu, H.; Wang, J.; and Jordan, M.~I.
\newblock 2017.
\newblock Deep transfer learning with joint adaptation networks.
\newblock In {\em ICML}.

\bibitem[\protect\citeauthoryear{Long \bgroup et al\mbox.\egroup
  }{2018}]{long2018conditional}
Long, M.; Cao, Z.; Wang, J.; and Jordan, M.~I.
\newblock 2018.
\newblock Conditional adversarial domain adaptation.
\newblock In {\em NeurIPS}.

\bibitem[\protect\citeauthoryear{Massa and Girshick}{2018}]{massa2018mrcnn}
Massa, F., and Girshick, R.
\newblock 2018.
\newblock {maskrcnn-benchmark: Fast, modular reference implementation of
  Instance Segmentation and Object Detection algorithms in PyTorch}.
\newblock \url{https://github.com/facebookresearch/maskrcnn-benchmark}.

\bibitem[\protect\citeauthoryear{Motiian \bgroup et al\mbox.\egroup
  }{2017}]{motiian2017few}
Motiian, S.; Jones, Q.; Iranmanesh, S.; and Doretto, G.
\newblock 2017.
\newblock Few-shot adversarial domain adaptation.
\newblock In {\em NeurIPS}.

\bibitem[\protect\citeauthoryear{Oh~Song \bgroup et al\mbox.\egroup
  }{2016}]{oh2016deep}
Oh~Song, H.; Xiang, Y.; Jegelka, S.; and Savarese, S.
\newblock 2016.
\newblock Deep metric learning via lifted structured feature embedding.
\newblock In {\em CVPR}.

\bibitem[\protect\citeauthoryear{Ren \bgroup et al\mbox.\egroup
  }{2015}]{ren2015faster}
Ren, S.; He, K.; Girshick, R.; and Sun, J.
\newblock 2015.
\newblock Faster r-cnn: Towards real-time object detection with region proposal
  networks.
\newblock In {\em NeurIPS}.

\bibitem[\protect\citeauthoryear{RoyChowdhury \bgroup et al\mbox.\egroup
  }{2019}]{roychowdhury2019automatic}
RoyChowdhury, A.; Chakrabarty, P.; Singh, A.; Jin, S.; Jiang, H.; Cao, L.; and
  Learned-Miller, E.
\newblock 2019.
\newblock Automatic adaptation of object detectors to new domains using
  self-training.
\newblock {\em arXiv preprint arXiv:1904.07305}.

\bibitem[\protect\citeauthoryear{Saito \bgroup et al\mbox.\egroup
  }{2017}]{saito2017maximum}
Saito, K.; Watanabe, K.; Ushiku, Y.; and Harada, T.
\newblock 2017.
\newblock Maximum classifier discrepancy for unsupervised domain adaptation.
\newblock {\em arXiv preprint arXiv:1712.02560}.

\bibitem[\protect\citeauthoryear{Saito \bgroup et al\mbox.\egroup
  }{2019}]{Saito2018Strong}
Saito, K.; Ushiku, Y.; Harada, T.; and Saenko, K.
\newblock 2019.
\newblock Strong-weak distribution alignment for adaptive object detection.
\newblock In {\em CVPR}.

\bibitem[\protect\citeauthoryear{Saito, Ushiku, and
  Harada}{2017}]{saito2017asymmetric}
Saito, K.; Ushiku, Y.; and Harada, T.
\newblock 2017.
\newblock Asymmetric tri-training for unsupervised domain adaptation.
\newblock In {\em ICML}.

\bibitem[\protect\citeauthoryear{Sakaridis, Dai, and
  Van~Gool}{2018}]{sakaridis2018semantic}
Sakaridis, C.; Dai, D.; and Van~Gool, L.
\newblock 2018.
\newblock Semantic foggy scene understanding with synthetic data.
\newblock {\em IJCV}.

\bibitem[\protect\citeauthoryear{Schroff, Kalenichenko, and
  Philbin}{2015}]{schroff2015facenet}
Schroff, F.; Kalenichenko, D.; and Philbin, J.
\newblock 2015.
\newblock Facenet: A unified embedding for face recognition and clustering.
\newblock In {\em CVPR}.

\bibitem[\protect\citeauthoryear{Shan, Lu, and Chew}{2018}]{shan2018pixel}
Shan, Y.; Lu, W.~F.; and Chew, C.~M.
\newblock 2018.
\newblock Pixel and feature level based domain adaption for object detection in
  autonomous driving.
\newblock {\em arXiv preprint arXiv:1810.00345}.

\bibitem[\protect\citeauthoryear{Shu \bgroup et al\mbox.\egroup
  }{2018}]{shu2018dirt}
Shu, R.; Bui, H.~H.; Narui, H.; and Ermon, S.
\newblock 2018.
\newblock A dirt-t approach to unsupervised domain adaptation.
\newblock In {\em ICLR}.

\bibitem[\protect\citeauthoryear{Simonyan and
  Zisserman}{2015}]{simonyan2014very}
Simonyan, K., and Zisserman, A.
\newblock 2015.
\newblock Very deep convolutional networks for large-scale image recognition.
\newblock In {\em ICLR}.

\bibitem[\protect\citeauthoryear{Tzeng \bgroup et al\mbox.\egroup
  }{2017}]{tzeng2017adversarial}
Tzeng, E.; Hoffman, J.; Saenko, K.; and Darrell, T.
\newblock 2017.
\newblock Adversarial discriminative domain adaptation.
\newblock In {\em CVPR}.

\bibitem[\protect\citeauthoryear{Wang \bgroup et al\mbox.\egroup
  }{2019a}]{wang2019few}
Wang, T.; Zhang, X.; Yuan, L.; and Feng, J.
\newblock 2019a.
\newblock Few-shot adaptive faster r-cnn.
\newblock {\em arXiv preprint arXiv:1903.09372}.

\bibitem[\protect\citeauthoryear{Wang \bgroup et al\mbox.\egroup
  }{2019b}]{wang2019multi}
Wang, X.; Han, X.; Huang, W.; Dong, D.; and Scott, M.~R.
\newblock 2019b.
\newblock Multi-similarity loss with general pair weighting for deep metric
  learning.
\newblock In {\em CVPR}.

\bibitem[\protect\citeauthoryear{Wu \bgroup et al\mbox.\egroup
  }{2018}]{wu2018dcan}
Wu, Z.; Han, X.; Lin, Y.-L.; Uzunbas, M.~G.; Goldstein, T.; Lim, S.~N.; and
  Davis, L.~S.
\newblock 2018.
\newblock Dcan: Dual channel-wise alignment networks for unsupervised scene
  adaptation.
\newblock In {\em ECCV}.

\bibitem[\protect\citeauthoryear{Yang \bgroup et al\mbox.\egroup
  }{2018}]{Yang2019Learning}
Yang, H.; Huang, D.; Wang, Y.; and Jain, A.~K.
\newblock 2018.
\newblock Learning continuous face age progression: A pyramid of gans.
\newblock In {\em CVPR}.

\bibitem[\protect\citeauthoryear{Zhang, David, and
  Gong}{2017}]{zhang2017curriculum}
Zhang, Y.; David, P.; and Gong, B.
\newblock 2017.
\newblock Curriculum domain adaptation for semantic segmentation of urban
  scenes.
\newblock In {\em ICCV}.

\bibitem[\protect\citeauthoryear{Zhang, Xie, and
  Lin}{2018}]{Zhang2018Photographic}
Zhang, Z.; Xie, Y.; and Lin, Y.
\newblock 2018.
\newblock Photographic text-to-image synthesis with a hierarchically-nested
  adversarial network.
\newblock In {\em CVPR}.

\bibitem[\protect\citeauthoryear{Zhu \bgroup et al\mbox.\egroup
  }{2017}]{zhu2017unpaired}
Zhu, J.-Y.; Park, T.; Isola, P.; and Efros, A.~A.
\newblock 2017.
\newblock Unpaired image-to-image translation using cycle-consistent
  adversarial networks.
\newblock In {\em ICCV}.

\bibitem[\protect\citeauthoryear{Zhu \bgroup et al\mbox.\egroup
  }{2019}]{zhu2019adapting}
Zhu, X.; Pang, J.; Yang, C.; Shi, J.; and Lin, D.
\newblock 2019.
\newblock Adapting object detectors via selective cross-domain alignment.
\newblock In {\em CVPR}.

\end{thebibliography}
\end{small}

\end{document}